\documentclass{amsart}
\usepackage[dvips]{graphicx}
\usepackage{amsmath}
\begin{document}

\title{Temporal plannability by variance of the episode length}
\author{B\'alint Tak\'acs, Istv\'an Szita and Andr\'as L\H{o}rincz}
\address{Department of Information Systems, E\"{o}tv\"{o}s Lor\'{a}nd University, P\'{a}zm\'{a}ny P\'{e}ter s\'{e}t\'{a}ny 1/C, Budapest, Hungary H-1117}

\begin{abstract}
Optimization of decision problems in stochastic environments is
usually concerned with maximizing the probability of achieving the
goal and minimizing the expected episode length. For interacting
agents in time-critical applications, learning of the possibility
of scheduling of subtasks (events) or the full task is an
additional relevant issue. Besides, there exist highly stochastic
problems where the actual trajectories show great variety from
episode to episode, but completing the task takes almost the same
amount of time. The identification of sub-problems of this nature
may promote e.g., planning, scheduling \textit{and} segmenting
Markov decision processes. In this work, formulae for the average
duration as well as the standard deviation of the duration of
events are derived. The emerging Bellman-type equation is a simple
extension of Sobel's work (1982). Methods of dynamic programming
as well as methods of reinforcement learning can be applied for
our extension. Computer demonstration on a toy problem serve to
highlight the principle.
\end{abstract}

\maketitle

\section{Motivation}

There is an increasing interest in planning in stochastic
environments, also called decision-theoretic planning. Since
planning is admired as a mainstay of artificial intelligence,
attempts were made to extend classical AI planning methods to
handle uncertainty. On the other hand, operations research has
made important advances recently in  solving decision making
problems \cite{Dyna91Sutton}. One of the earliest works, which
tried to unify the two complementary approaches appeared in the
seventies \cite{Feldman77Decision}. Recently Markov decision
processes are proposed as a unifying framework for
decision-theoretic planning
\cite{Boutilier99Decision,Dean95Planning}. The success of the
closely related reinforcement learning (RL) techniques have
encouraged work in planning using Markov decision processes.

For planning one needs goals.\footnote{Note that planning in the
field of reinforcement learning is sometimes used in the context
of off-line optimization of decision making (\cite{Dyna91Sutton}).
Here, planning is used in its ordinary meaning, i.e., as a synonym
for considering different trajectories to achieve the goal, or for
scheduling.} Classical AI planning usually determines the goal as
a subset of the state space. This definition is extended by Markov
decision processes, where an immediate cost (or reward) function
is defined and the `goal' of the decision maker is not to reach a
set of states, but to minimize the discounted long-term cumulated
cost. In RL, planning is equal to finding `good' policies, i.e.
state-dependent action selection strategies for which the
cumulated cost is minimal \cite{Littman98Computational}.
\label{plan_strict} In Markov decision processes, one speaks of
episodic tasks, when well-defined terminal (goal) states exist and
the episode as well as the accumulation of costs stops in finite
time --- after reaching one of the terminal states. In episodic
tasks, planning is often formalized as the task of finding plans
for which the probability of achieving the goal states is maximal
and/or the expected episode length is minimal.

The suggested plans may have many evaluation criteria, such as the
average cost accumulated during execution, the probability of a
successful execution, or the expected episode length. We will
focus here on a rarely emphasized feature of plans, namely the
\emph{reliability of the plan execution time}, which is an
important issue in time-sensitive systems. For example, imagine a
cooperating multi-agent system, where the agents can interact only
if they are close in space (e.g. the agents are the trucks of a
transport company). It is impossible to create a global long-term
plan if the arrival time of specific agents (trucks) show a great
variety. Being accurate in the time of arrival may be much more
important than having somewhat faster deliveries on the average.
Another example is that `the postman may bring a letter at 8
o'clock'. It might be important to check the mailbox, although the
probability of receiving an answer is low. Missing an appointment
may cause long-term effects in the whole system like a small
snowball can grow into an avalanche. This is a relevant issue in
every system with strongly nonlinear responses, which is the
common cause in the engineering practice. In most real-life
problems, for long-term planning nearly-deterministic (i.e.,
\emph{reliable}) sub-components (sub-tasks) are necessary. A
number of possible applications are mentioned at the end of the
paper.

How to measure reliability? The simplest assumption is that the
larger the variance of the duration of a sub-task, the less one
can rely on that component when making a plan. In the next section
we propose a simple algorithm for calculating the time variance of
episodes.

\section{Calculating the duration and the variance of the duration of episodes}\label{s_calc}

\subsection{Assumptions and definitions}

Consider an episodic MDP with a finite state space $S$ and a
finite action space $A$. The agent starts from some state $x_0$,
and makes steps according to policy $\pi: S \times A \to [0,1]$
until it reaches a state in the terminal set $Z \subset S$. Let
$Z_0\subset Z$ be the goal set. We may assume that from any
terminal state $z \in Z$, the agent is transferred to a
hyper-terminal state $\omega$, i.e. $P(z,a,\omega)$=1 for all
$z\in Z, a\in A$. This assumption does not modify the time of
reaching $Z$ for the first time, but it enables us to simplify our
formalism, because every state in $Z$ is visited at most once.

Denote by $p^{\pi}(x,y)$ the probability that from state $x$ the
agent arrives to state $y$ while following policy $\pi$, i.e.

\begin{eqnarray}
p^{\pi}(x,y) =\sum_{a\in A} \pi(x,a) P(x,a,y)\label{mc_simp}.
\end{eqnarray}

The visited states in a start-goal episode are noted by $\left\{
b_0,\ldots,b_K\right\}$ assuming the episode takes $K$ steps (note
that $K$ is also a random variable). If an $x \rightarrow y$
transition takes $\tau_{xy} \in \mathcal{N}_0$ time,\footnote{The
time spent by a transition could depend also on the applied
action, which constrains one to work with function $\tau(x,a,y)$
instead of $\tau_{xy}$. The emerging equations have similar and
are omitted here for the sake of simplicity. The only difference
is that the simplified notation of Eq.~(\ref{mc_simp}) can not be
used in the more general case.} then the completion time of an
episode can be calculated as $T=\sum_{i=0}^{K-1}\tau_{b_i
b_{i+1}}$ where $T \in \mathcal{N}_0$. Naturally, if every
transition takes 1 unit of time then $K \equiv T$.

We would like to find the answer to the following three questions:
\begin{enumerate}
\item What is the probability that the agent ends up in a \emph{goal state}, i.e. in $Z_0$?
\item What is the average time needed for reaching a goal state?
\item What is its variance?
\end{enumerate}

\subsection{The probability of success}

An episode may be either successful (the agent ends up in $Z_0$)
or unsuccessful (the agent ends up in $Z \backslash Z_0$). Denote
by $s(x)$ the probability that starting from $x$, the agent will
be successful, i.e. $s(x) = \Pr(b_K \in Z_0| b_0 = x)$. Clearly,
\begin{eqnarray}
s(x) =
  \begin{cases}
    1 & \text{if $x\in Z_0$}, \\
    0 & \text{if $x\in Z\backslash Z_0$}, \\
    \sum_{y\in S} p^{\pi}(x,y) s(y) & \text{if $x\in S \backslash Z$}.
  \end{cases}\label{eq:S_value}
\end{eqnarray}

\subsection{The probability of success in exactly T time}

Let $q(T|x)$ denote the probability of reaching $Z_0$ exactly at
time $T$, assuming the agent started from state $x$ at time $0$.
That is, $q(T|x) = \Pr(T, b_K \in Z_0|b_0 = x)$ for every $T \ge
0$. Then,
$$
  q(0|x) = \begin{cases}
    1 & \text{if $x\in Z_0$}, \\
    0 & \text{if $x\in S\backslash Z_0$},
  \end{cases}
$$
and the following simple recursion holds:
$$
  q(T|x) = \begin{cases}
   \sum_{y\in S} p^{\pi}(x,y) q(T-\tau_{xy}|y) & \text{if $x \in S \backslash Z$ } \\
   0 & \text{if $x \in Z$}
  \end{cases}
$$
for $T \ge 1$ and $q(T|x)=0$ for every $T<0$.

For the sake of simplicity, we assume that from any non-terminal
state the agent reaches a terminal state in finite time with
probability one, i.e.
$$\{ \Pr(b_K \notin Z_0|b_0 = x), q(0|x), q(1|x), q(2|x), \ldots \}$$
is a probability distribution. It is easy to see that
$\sum_{T=0}^\infty q(T|x) = s(x)$.

\subsection{The average episode length}

Making use of the above recursion, similar recursion can be
derived for the expected number of time steps needed to reach a
goal state from $x$ (denoted by $A(x)$):

\begin{eqnarray*}
    A(x) &=& \mathbf{E} (T|b_K \in Z_0, b_0 = x) = \sum_{T=0}^\infty T \cdot \Pr(T|b_K \in Z_0, b_0 = x) = \\
         &=& \sum_{T=0}^\infty T \cdot \frac{\Pr(T, b_K \in Z_0|b_0 = x)}{\Pr(b_K \in Z_0|b_0 = x)} = \frac{1}{s(x)}\sum_{T=0}^\infty T \cdot
         q(T|x).
\end{eqnarray*}
In particular, if $x\in Z$, then $A(x)=0$. For $x\in S\backslash
Z$, $A(x)$ can be expressed as
\begin{eqnarray}
  A(x) &=& \frac{1}{s(x)} \sum_{T=0}^\infty T \cdot q(T|x) =  \nonumber\\
       &=& \frac{1}{s(x)} \sum_{T=0}^\infty T \sum_{y\in S} p^{\pi}(x,y) q(T-\tau_{xy}|y) = \nonumber\\
       &=& \frac{1}{s(x)} \sum_{y\in S} p^{\pi}(x,y) \sum_{T=0}^\infty \bigl[ (T-\tau_{xy}) \cdot q(T-\tau_{xy}|y) + \tau_{xy} q(T-\tau_{xy}|y)\bigr] = \nonumber\\
       &=& \frac{1}{s(x)} \sum_{y\in S} p^{\pi}(x,y) \left( \sum_{T=0}^\infty T \cdot q(T|y) + \tau_{xy} \sum_{T=0}^\infty q(T|y)\right) = \nonumber\\
       &=& \frac{1}{s(x)} \sum_{y\in S} p^{\pi}(x,y) s(y) \left[ A(y) +
       \tau_{xy}\right]\label{eq:A_value}.
\end{eqnarray}

\subsection{The variance of episode length}

The second moment $B(x)$ of the episode length for each $x\in S$
can be derived in a similar manner:
$$
  B(x) = \mathbf{E} (T^2|b_K \in Z_0, b_0 = x) = \frac{1}{s(x)}\sum_{T=0}^\infty T^2 \cdot q(T|x).
$$
If $x$ is a terminal state, then $B(x)=0$. Otherwise,
\begin{eqnarray}
  B(x) &=& \frac{1}{s(x)} \sum_{T=0}^\infty T^2 \cdot q(T|x) = \frac{1}{s(x)} \sum_{T=0}^\infty T^2 \sum_{y\in S} p^{\pi}(x,y) q(T-\tau_{xy}|y) = \nonumber\\
       &=& \frac{1}{s(x)} \sum_{y\in S} p^{\pi}(x,y) \cdot \sum_{T=0}^\infty \bigl[ (T-\tau_{xy})^2 \cdot q(T-\tau_{xy}|y) + \nonumber\\
       &+& 2 \cdot \tau_{xy}\cdot T \cdot q(T-\tau_{xy}|y) - \tau_{xy}^2 q(T-\tau_{xy}|y) \bigr] = \nonumber\\
       &=& \frac{1}{s(x)} \sum_{y\in S} p^{\pi}(x,y) \cdot \sum_{T=0}^\infty \bigl[ (T-\tau_{xy})^2 \cdot q(T-\tau_{xy}|y) + \nonumber\\
       &+& 2 \cdot \tau_{xy}\left[ (T-\tau_{xy}) \cdot q(T-\tau_{xy}|y) + \tau_{xy} \cdot q(T-\tau_{xy}|y) \right] - \nonumber\\
       &-& \tau_{xy}^2 q(T-\tau_{xy}|y) \bigr] = \frac{1}{s(x)} \sum_{y\in S} p^{\pi}(x,y) \cdot \nonumber\\
       &\cdot& \left(\sum_{T=0}^\infty T^2 \cdot q(T|y) + 2 \cdot \tau_{xy} \sum_{T=0}^\infty T \cdot q(T|y) + \tau_{xy}^2 \sum_{T=0}^\infty q(T|y) \right) = \nonumber\\
       &=& \frac{1}{s(x)} \sum_{y\in S} p^{\pi}(x,y) s(y) \left[ B(y) + 2 \cdot \tau_{xy} \cdot A(y) + \tau_{xy}^2
       \right]\label{eq:B_value}.
\end{eqnarray}

By the well known formula, the variance is
\begin{eqnarray}
D(x) = \sqrt{B(x) - A(x)^2}\label{eq:C_value}.
\end{eqnarray}

These recursive equations form the base of an algorithm: first,
iterate the $s(x)$ quantity according to Eq.~(\ref{eq:S_value}),
until equilibrum is reached, second, iterate $A(x)$ using
Eq.~(\ref{eq:A_value}) with the previously calculated $s(x)$,
third, iterate $B(x)$ with Eq.~(\ref{eq:B_value}) using $s(x)$ and
$A(x)$. The resulting method gives (i) the probability of success,
(ii) the time an episode takes on average, and (iii) the variance
of the episode time from all possible starting states for a
prescribed terminal state set.

\section{Demonstration}\label{s_dem}

\begin{figure}
\centering
\includegraphics[width=10cm]{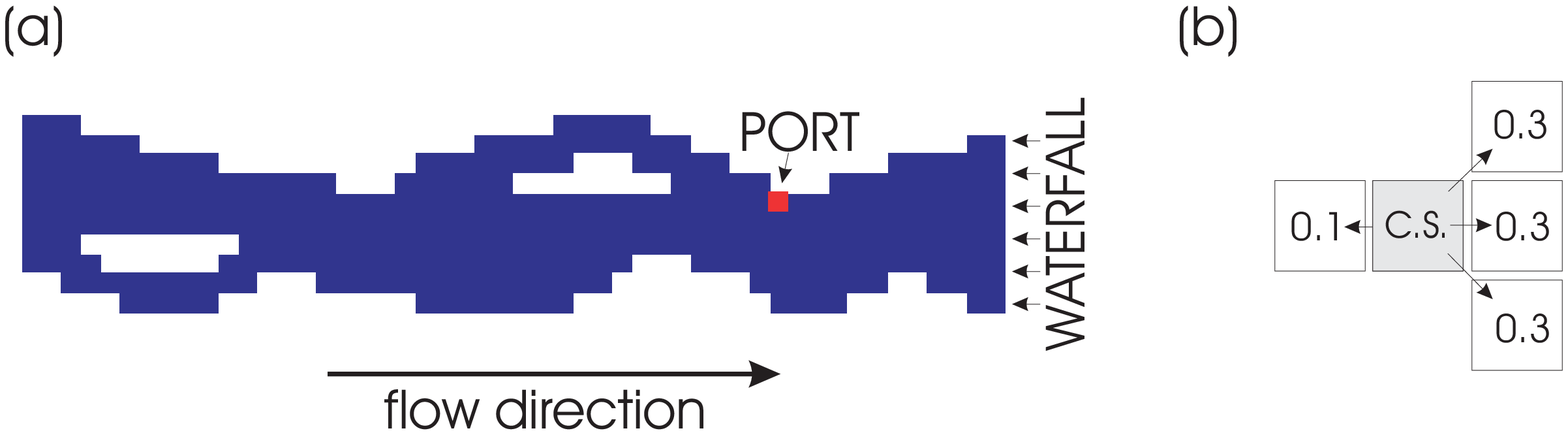}
  \caption{\textbf{Description of the toy problem.}
  \newline Subfigure (a) shows the `river' flowing from the left to the right.
  The river is constructed on a 50 by 10 square grid. The `port' is marked with a red square.
  The river empties into a waterfall on the right
  side. If the ship reaches any states of the last column, the trial has an
  unsuccessful end. Subfigure (b) shows the predefined state
  transitions. If not obstructed by the river's bank, there is an equal
  probability (0.3) of transferring from the current state (noted by C.S.) into one of
  the three neighboring states in the next right column. The diagonal steps take 2 time units,
  while the forward step takes 1 time unit. Simultaneously, there is also a low probability (0.1) of
  moving backwards to the left column, which move takes 5 time units. The banks of the river and the
  islands limit the number of available transitions. In this cases, the probability of the unavailable transition(s)
  is equally distributed between the remaining ones.
  }\label{fig:flow}
\end{figure}

\begin{figure}
\centering
\includegraphics[width=10cm]{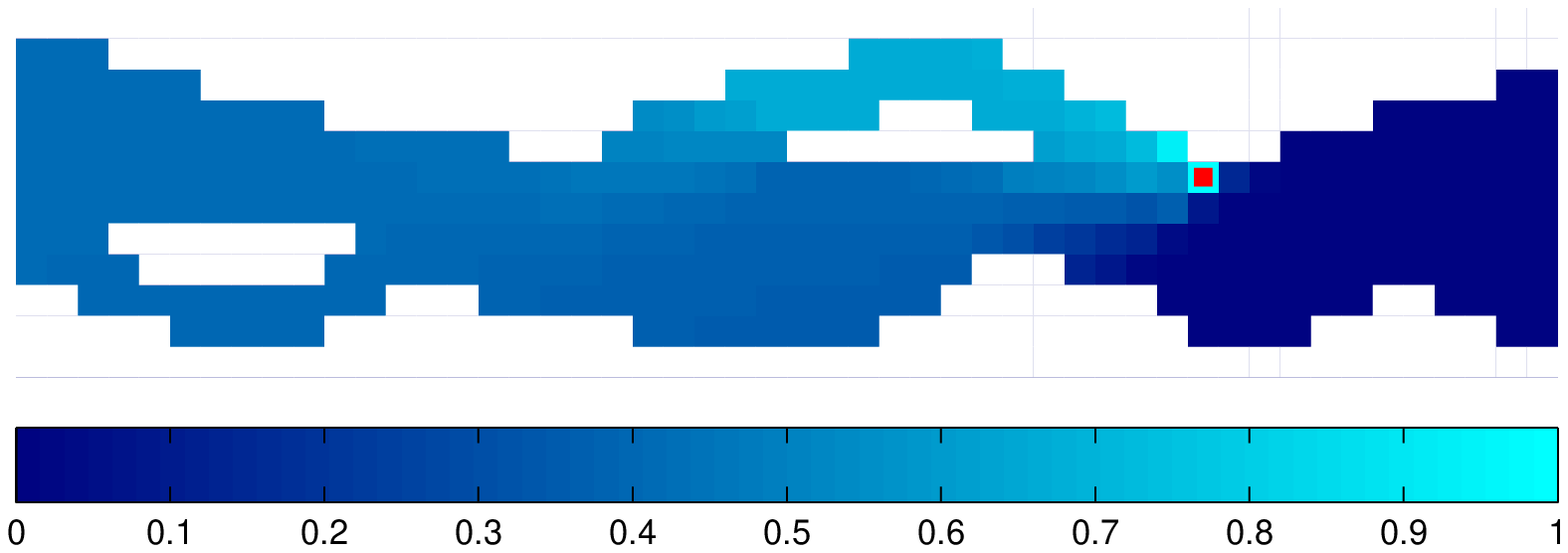}
  \caption{\textbf{Probability of a successful episode.}
  \newline The figure plots the probability of successful arrivals from any
  given state ($s(x)$) (computed by the recursive formula (\ref{eq:S_value}).
  The graded blue color of a point of the river provides the probability of arriving to the port.
  Color coding is provided at the bottom of the figure. The dark colors on the right hand
  side of the port indicate that chances of successful arrivals from that part of the
  river are low.
  }\label{fig:S_value}
\end{figure}

\begin{figure}
\centering
\includegraphics[width=10cm]{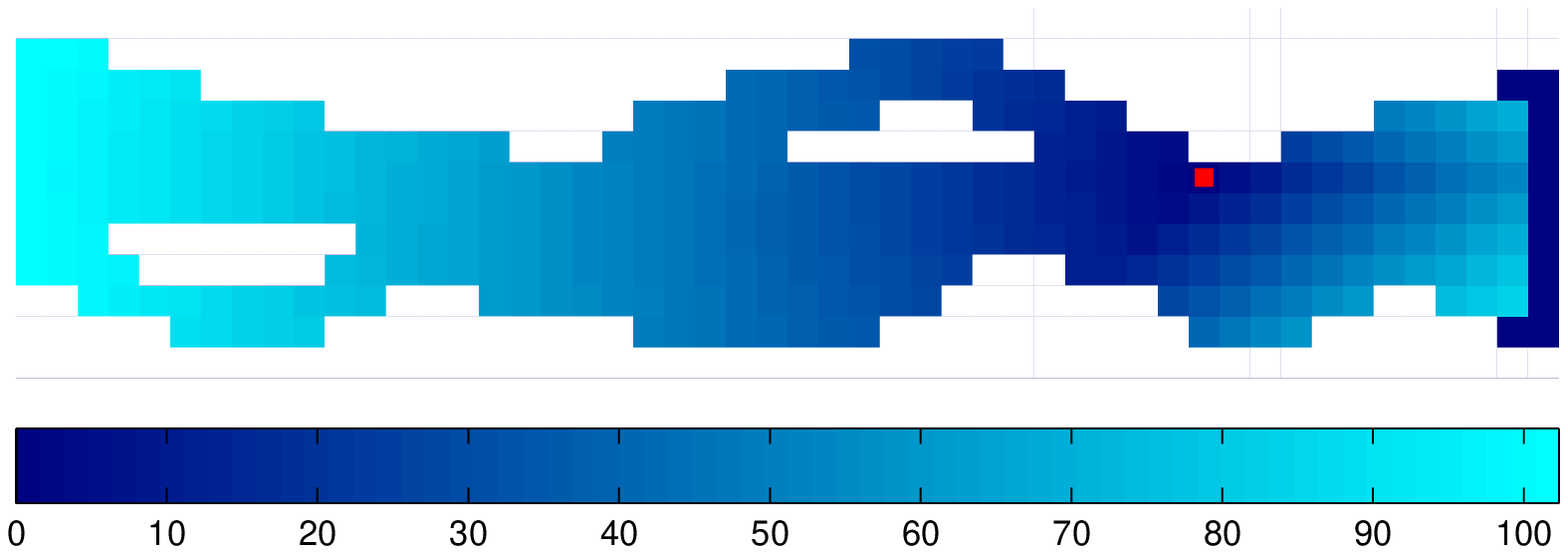}
  \caption{\textbf{Average episode lengths.}
  \newline Average durations ($A(x)$) of successful episodes are indicated by graded blue color for every state ($x$).
  }\label{fig:A_value}
\end{figure}

\begin{figure}
\centering
\includegraphics[width=10cm]{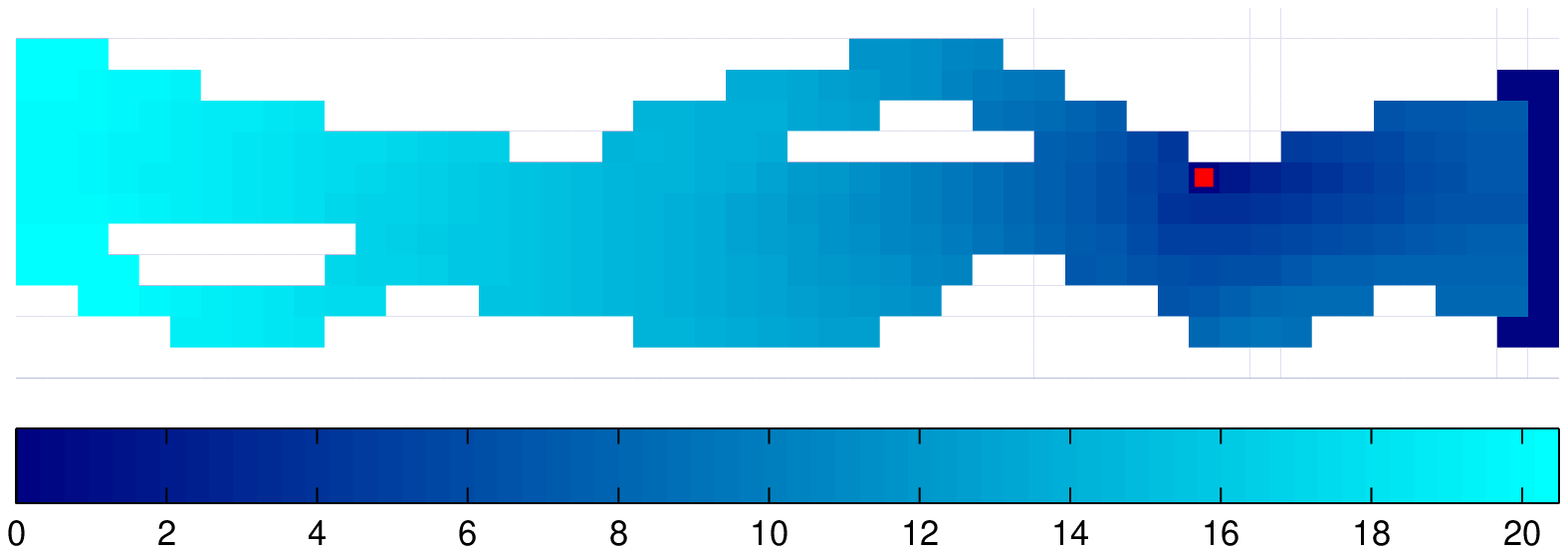}
  \caption{\textbf{Variance of episode length.}
  \newline The variance of the duration of successful episodes ($C(x)$) are indicated by graded blue color for every state $(x)$.
  }\label{fig:C_value}
\end{figure}

We generated a simple toy problem to demonstrate the utility of
the algorithm. Imagine a river, which flows rapidly from west to
east. There is a port on the left bank of the river. There is also
a ship on the river, initially positioned somewhere on the river.
The ship has to reach the port, otherwise it enters a terrible and
huge waterfall, situated at the east end of the river. The water
has very strong vortices and currents, therefore it is almost
impossible to control the movement of the ship. Our model assumes
that in the next time step the currents will almost surely drive
the ship towards the waterfall, but we don't know exactly in which
direction. A small chance exists that the vortices drive the ship
backwards, too. Of course, at a given instant, and if not at the
east end of the river, the ship can't be very far from its
previous position. Details of this toy problem are depicted in
Fig. \ref{fig:flow}. This example has a few situations, which will
occur with high probabilities: every transition (except for the
ones near the banks) has low probability to be happened. Direct
planning of particular trajectories is almost meaningless. But, as
it can be seen on Figs.~\ref{fig:A_value} and \ref{fig:C_value},
temporal planning, in the sense defined in Section~\ref{s_calc},
is still meaningful. Figures \ref{fig:S_value}, \ref{fig:A_value}
and \ref{fig:C_value} plot the quantities $s(x)$, $A(x)$ and
$C(x)$ respectively, after reaching equilibrium (we performed 100
iterations). The figures show that there is a very low chance to
reach the port when the ship starts behind it, as expected.
However, because we calculate the time needed for
\emph{successful} episodes, the average episode times needed are
almost the same as when the ship starts in the leftmost positions,
but the reliability of these times -- measured by the variance
(Fig. \ref{fig:C_value})  -- are higher in the first case. If the
harbor master knows the ship's starting position, than he knows
when to look for the ship on the horizon according to Fig.
\ref{fig:A_value}, and he can decide from Fig. \ref{fig:C_value}
how long should he wait until he can almost surely guess that the
ship is lost.

\section{Discussion}

Examining the probabilistic properties of the cumulated cost
beyond its average value is not new in the literature. The
attempts concentrated mainly on penalizing the variance of the
accumulated cost
\cite{Filar89Variance,Greffenstette90Learning,Huang94Finding,White94Computational,White94Mathematical,Collins94Finite}.
Algorithms for calculating the variance directly have also been
published. Kemeny and Snell (\cite{Kemeny60Finite}) developed a
formula for the second moment of first-passage times in Markov
chains. Formulae are derived in \cite{Platzman78Mimeographed} for
the second moment of the accumulated total reward of Markov chains
with rewards. In \cite{Sobel82Variance} Sobel presented a general
method for calculating the arbitrary moments of the cumulated
discounted cost. This method is similar to ours. The algorithm
introduced here emphasizes the possibility of using the average
duration of the episode, or higher moments of the duration of an
episode as costs to transfer the problem of plannability to the
domain of reinforcement learning, as it will be discussed shortly.
By introducing the probability of success, a novel property of the
formulae is that they can be used to compute the appropriate
properties for successful or unsuccessful episodes.

\subsection{Reinforcement Learning Methods}\label{s_rl}

The algorithm does not employ the immediate costs (rewards)
defined by the original MDP. Nevertheless, the algorithm is
closely related to such cost-based approaches: If the transition
time is viewed as cost, then calculating the average time of the
execution of the episode can be seen as calculating the average
sum of the (time-)costs in the non-discounted case. In this sense,
the quantities $\tau_{xy}$ determine the corresponding cost
function. Indeed, we can recognize Eq.~\ref{eq:B_value} as the
well-known Bellman-equality \cite{Bellman57Dynamic} for the case
of non-discounted costs.

Therefore, similarly to the standard reinforcement learning
problem, we do not need to solve the DP equations directly, but we
may update the $s$, $A$ and $B$ values in different ways explored
by theoretical works in the field of RL (for a historical review,
see, e.g., \cite{Sutton98Reinforcement} and references therein).
Asynchronous and Monte Carlo methods, as well as sampling methods
analogous to Q-learning or SARSA can be of use here. The advantage
of RL methods as compared to DP methods can be seen in problems
that are too large for direct computations. Moreover, RL methods
can be used by direct interaction with the system (called on-line
learning). This solution is the only option when no model is
available. On the other hand, if a DP model is available and
sampling methods with the real system are cumbersome, then DP has
the advantage of being an off-line method, not limited by
constraints on interactions with a real system.

Another advantage of formulating temporal planning within the
framework of RL is that costs on time (i.e., planning) and real
costs of the episode can be unified into a single cost function.
Moreover, the multi-criteria paradigm
\cite{Mitten64Composition,gabor98multi-criteria} can be applied to
consider multiple cost functions in parallel. In particular, cost
functions with mean-variance tradeoffs
\cite{Filar89Variance,Huang94Finding,White94Computational,White94Mathematical,Collins94Finite}
may penalize solutions with high variance. One has the option of
solving the MDP for the original cost function and also for the
time as the `cost' and then combine the two aspects.

\subsection{Planning as a Segmentation Tool}

As it has been mentioned before, in our terminology planning is an
off-line method, which serves to minimize the frequency of on-line
decision making. Planning can be the tool to segment decision
making problems into subproblems.

In practice, the major problem of decision-theoretic planning is
the intractably large space state. Handling complex, real-life
situations is impossible without grouping states into larger sets
(possibly structured in hierarchical fashion), thus allowing for
crude discretization, partial observation and assigning
\emph{subgoals} for complex tasks. In most real-life problems, the
environment is only partially observable, therefore we need to
solve partially observable Markov decision processes instead of
classical MDPs. Partial observability may come from the limited
perceptual capabilities of the agent: the agent may possess
insufficient resources to observe all variables of its
environment, the agent may be localized not having information
about remote objects in space, the agent may be slow to observe
all temporal changes happening in the environment. If the
optimization problem is partially observable, then solving this
problem is intractable in general \cite{Littman96Algorithms}.
Every available tool should be used to ease the learning problem.
The most promising methods are those, which are only partly
limited by the number of interactions with the real system, that
is, the on-line methods extended by off-line methods. Off-line
computations are preferred when they are relatively inexpensive.
These methods may be able to segment the current problem into
subproblems. One of the candidates for finding a suitable
segmentation is the temporal reliability measured by the variance
of the duration of the episode or a part of the episode. One can
use the variance calculated by Eq.~(\ref{eq:C_value}) to measure
which states are reliable subgoals. Then, variance of the duration
of the episode serves as auxiliary information for distinguishing
states which can be later assigned as subgoals. The advantages of
subgoals and concept formation using subgoals have been in the
focus of recent research interest (see, e.g.,
\cite{Ring91Incremental,Schmidhuber91Learning,Wiering96HQ,Sun00Self,McGovern02PHD}
and references therein).

These concepts may gain applications in a variety of areas:
basically everywhere, where RL and DP need to be extended by
scheduling. A few representative examples are as follows: job-shop
scheduling \cite{zhang95reinforcement}, elevator optimization
\cite{crites96improving}, robotic soccer \cite{kitano97robocup} or
context focused Internet crawling making use of RL
\cite{Kokai02Fast}.

\section{Acknowledgements}

This work was supported by the Hungarian National Science
Foundation (Grant OTKA 32487) and by EOARD (Grant
F61775-00-WE065). Any opinions, findings and conclusions or
recommendations expressed in this material are those of the
authors and do not necessarily reflect the views of the European
Office of Aerospace Research and Development, Air Force Office of
Scientific Research, Air Force Research Laboratory.

\bibliographystyle{amsplain}
\bibliography{temp_plan}

\end{document}